%% file: _main.tex
\input{_constants}
\arxiv % \review OR \arxiv OR \cameraready

\pdfoutput=1
\documentclass[10pt,twocolumn,letterpaper]{article}
\input{cvpr_header}

\unless\ifarxiv \myexternaldocument{_supplementary} \fi

\begin{document}
% %% TITLE
% \title{Dynamic-I2V: Exploring Image-to-Video Generaion Models via Multimodal LLM}
% \author{
%     Peng Liu\textsuperscript{1*}, Xiaoming Ren\textsuperscript{1*}, Fengkai Liu\textsuperscript{1,2*}, Qingsong Xie\textsuperscript{1*}, \\
%     Quanlong Zheng\textsuperscript{1}, Yanhao Zhang\textsuperscript{1†}, Haonan Lu\textsuperscript{1}, Yujiu Yang\textsuperscript{2} \\   
    
%     \textsuperscript{1}OPPO AI Center \\
%     \textsuperscript{2}Tsinghua University \\
%     \\ 
%     \footnotesize{* Equal contribution. † Corresponding Author}
% }

%% TITLE
\title{Dynamic-I2V: Exploring Image-to-Video Generation Models via Multimodal LLM}
\author{
    Peng Liu\textsuperscript{1*} \hspace{0.5cm} Xiaoming Ren\textsuperscript{1*}\hspace{0.5cm} Fengkai Liu\textsuperscript{1,2*}\hspace{0.5cm} Qingsong Xie\textsuperscript{1} \\
    Quanlong Zheng\textsuperscript{1}\hspace{0.5cm} Yanhao Zhang\textsuperscript{1†}\hspace{0.5cm} Haonan Lu\textsuperscript{1}\hspace{0.5cm} Yujiu Yang\textsuperscript{2} \\ 
    \textsuperscript{1}OPPO AI Center \hspace{1cm}
    \textsuperscript{2}Tsinghua University \\[5pt]
}

\makeatletter
\g@addto@macro\@maketitle{
    \vspace{-30pt}
        \begin{figure}[H]
          \setlength{\linewidth}{\textwidth}
          \setlength{\hsize}{\textwidth}    
          \centering
          \includegraphics[width=0.95\linewidth]{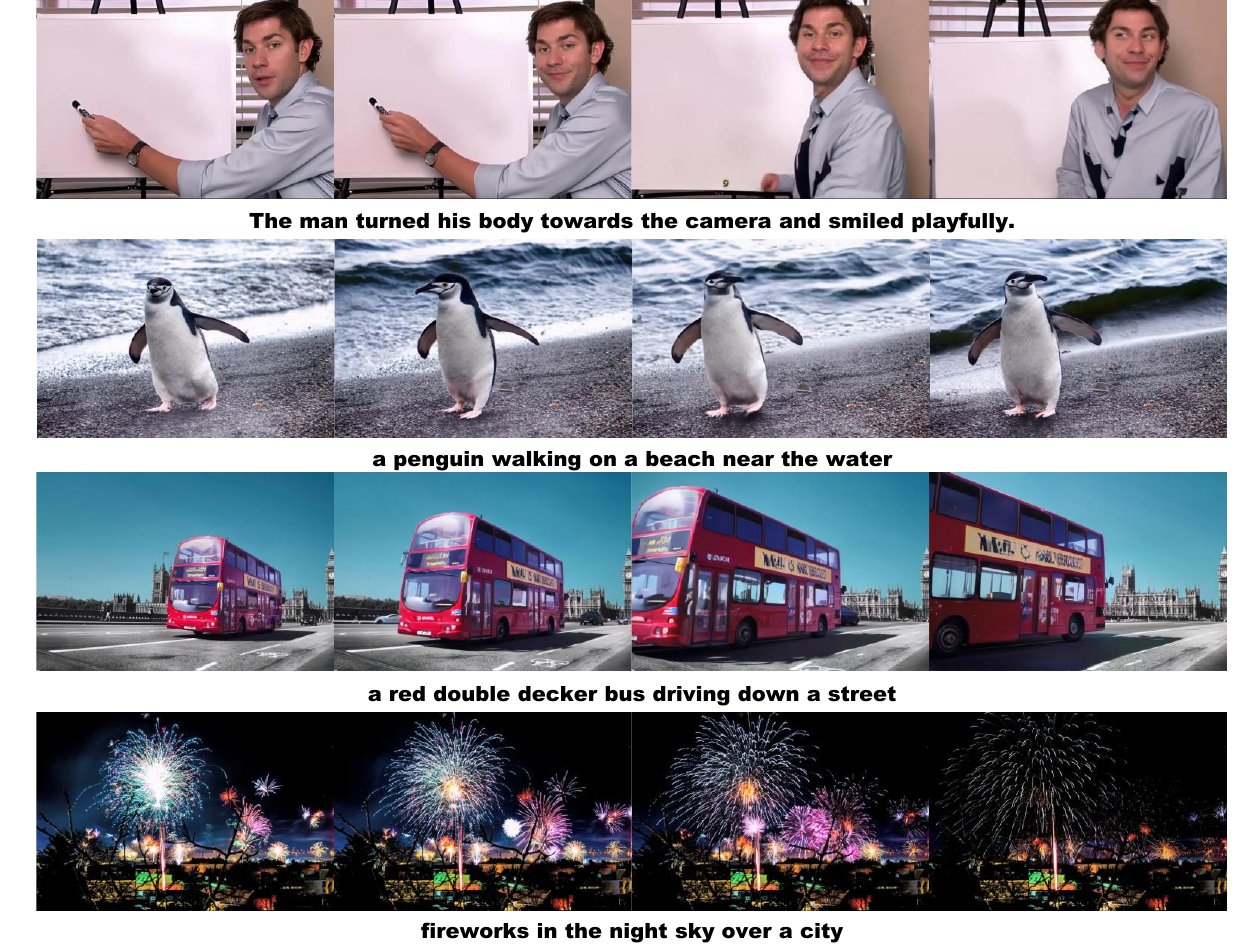}
          \caption{Some results generated by Dynamic-I2V. Our results exhibit high dynamics while ensuring the quality of generation.}
        \end{figure}
}

% \vspace{-30pt}
%   \begin{figure}[H]
%   \setlength{\linewidth}{\textwidth}
%   \setlength{\hsize}{\textwidth}
%     \centering
%     \includegraphics[width=0.95\linewidth]{figs/showcase4.pdf}
%   \caption{Some results generated by Dynamic-I2V. Our results exhibit high dynamics while ensuring the quality of generation.}
% \end{figure}

\maketitle

% \begin{minipage}{\linewidth}
%     \centering
%     \textbf{在图像上方的文字} % 在图像上面可以写一些文本
%     \vspace{0.5em}
%     \includegraphics[width=0.8\linewidth]{your-image-file.png}
%     \captionof{figure}{这是一个图像的说明}
%     \label{fig:my_label}
% \end{minipage}

% Home page Figure
% \vspace*{-10mm}
% \begin{figure*}[!ht]
%   \centering
%   \includegraphics[width=\textwidth]{figs/showcase4.pdf} % 图片路径
%   \caption{Some results generated by Dynamic-I2V. Our results exhibit high dynamics while ensuring the quality of generation.}
%   \label{fig:teaser}
%   \vspace{-5mm} % 调整图片与摘要的间距
% \end{figure*}

\renewcommand{\thefootnote}{} % 使脚注不显示任何编号
\footnotetext[1]{* Equal contribution.}  % 脚注1
\footnotetext[2]{† Corresponding Author.}  % 脚注2

\input{00_abstract}
\input{01_intro}
\input{02_related}

\input{03_method}
\input{04_eval}

\input{05_exp}

\input{10_conclusion}

{
    \small
    \bibliographystyle{ieeenat_fullname}
    \bibliography{11_references}
}

% \ifarxiv 
% \clearpage \appendix \input{12_appendix}  \fi

\end{document}

%% TITLE
\title{\paperTitle}
\author{\authorBlock}
\maketitlesupplementary

\appendix

\section{Human Evaluation}
\label{sec:appendix}

\subsection{Scoring Rules}

We recruited 20 volunteers to rank their preferences for videos generated by four distinct models across several evaluative dimensions. Subsequently, an overall quality preference ranking of the videos was also conducted. Scores were then assigned based on the degree of preference indicated by the volunteers.\\
Finally, the proportion of each model's score in the total score was calculated for each dimension.This approach facilitates an intuitive comparison of the popularity levels among the various models. We provided reference standards below for the volunteers to consult during their evaluation process.\\
Given that some videos may not exhibit discernible differences in quality within certain subjective dimensions, volunteers were permitted to abstain from responding to specific items. Consequently, the scoring protocol was adapted to accommodate such instances. The allocation of scores was adjusted based on the number of responses collected for each item, utilizing the following formula:
\[
\textrm{score} = \frac{\sum \textrm{Frequency}\times \textrm{Weight}}{\textrm{Total\ number\ of\ volunteers}}
\]
This approach ensures that the more respondents there are for a given question, the higher the total score allocated to that question, thereby indicating a greater level of confidence in the results of that particular item. However, it is important to note that the relative differences between the options within the same question remain unaffected, preserving the integrity of the comparative analysis among the choices provided.
\subsection{Reference Standards}
\begin{itemize}
    \item \textbf{Dynamics degree}: Mainly focus on the movement range of the video. The larger the movement range, the higher the ranking. It can be evaluated from the following dimensions:
        \begin{enumerate}
            \item \textbf{Frame consistency} means that there needs to be a certain continuity between frames. If there are no deformation artifacts, blurry or distorted objects, or objects that suddenly appear or disappear, then we think that the frame consistency is not good.
            \item \textbf{Completeness of movement and range of motion}: Mere camera movement does not count as achieved motion. A win in this dimension indicates greater motion, even if it includes distortion, is fast-moving, or looks unnatural. Greater range of motion and fullness of movement is desirable.
            \item \textbf{The naturalness of the movement}: Natural and realistic movements need to conform to the laws of nature and physics. For certain aspects, such as the character's body movements and facial expressions, they should appear natural under human subjective evaluation.
        \end{enumerate}
    \item \textbf{Natural degree}: The more a video resembles the real world, the better its quality and the higher it should be ranked. The more it resembles an AI-generated video, the lower it should be ranked.
    \item \textbf{Text compliance}: Whether the video can generate an accurate video based on the text provided. The main focus is on whether the model can recognize objects in the image and associate the image with the text.
    \item \textbf{Overall quality}: Rate the overall video quality based on personal preferences
\end{itemize}

\subsection{Detailed Results}
The scores and calculation process for 'Overall quality' are presented on Tab.~\ref{tab:t1}. The scores reported in the main text are normalized score.
% Table
\begin{table}[]\small
\begin{minipage}{0.45\textwidth}
\caption{Summary of human study results for Overall Quality}
\vspace{-10pt}
\begin{center}
\begin{tabular}{lllll}
  \toprule
    Models(anonymous) & M1 & M2 & M3 & M4 \\ \midrule
    Video 1   & 1.4   & 3.0 & 3.0 & 2.6  \\ 
    Video 2   & 2.4   & 3.2 & 3.4 & 1.0 \\ 
    Video 3   & 2.4   & 3.2 & 3.2 & 1.2 \\ 
    ...   & ...   & ... & ... & ... \\ 
    Overall Score  & 99.8   & 149.6 & 127.6 & 77  \\
    Normalized Score(\%) & 21.98   & 32.95 & 28.11 & 16.96\\
    Rank   & 3   & 1 & 2 & 4 \\
  \bottomrule
\end{tabular}
\end{center}
\label{tab:t1}
\end{minipage}
\end{table}%

\section{More visualization results}

In the main text, we presented two examples comparing Dynamic-I2V and CogVideoX-I2V-5B. Due to space constraints, we did not include the results of DynamiCrafter and SVD which are also used in the Human Study. Here, we will supplement the results of all these models.

Additionally, extracting frames from a video and displaying them often fails to effectively demonstrate the generated video's quality. On one hand, limited image resolution makes it difficult to showcase certain details. On the other hand, aspects like camera panning, especially in landscape scenes, are better conveyed through video format. Therefore, we have included an additional file, results.zip, which contains several results on VBench-I2V test set.

\begin{figure*}[] 
  \centering
  \includegraphics[width=0.95\textwidth]{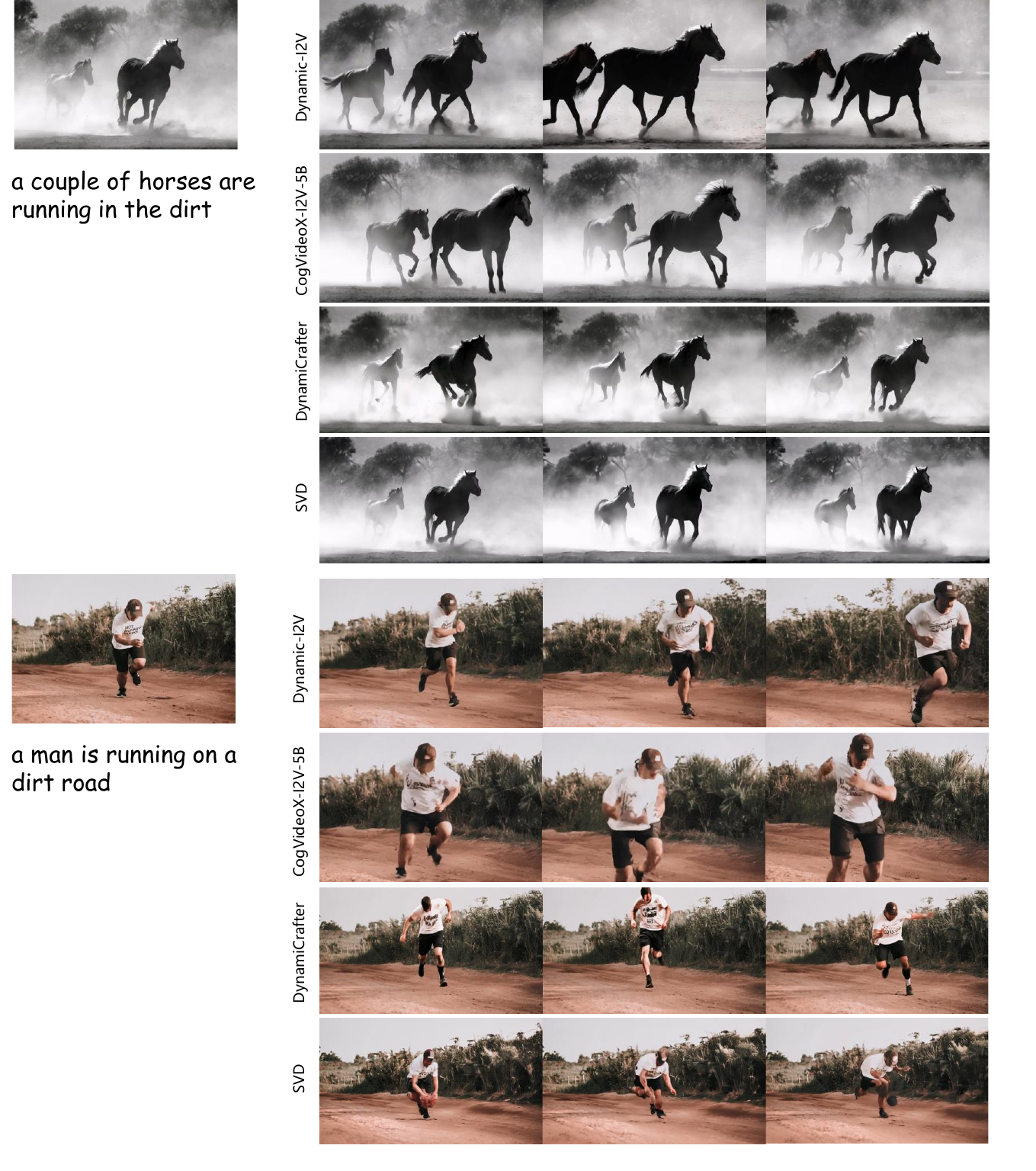}
  \caption{Subjective comparison results of Dynamic-I2V, CogVideoX-I2V-5B, DynamiCrafter and SVD. Note that SVD generated videos without any prompts.}
 \vspace{-5pt}
\label{fig:Examples}
\end{figure*}

\section{User prompt for MLLM}

\begin{table*}[]
\caption{User prompt for MLLM.}
\vspace{-10pt}
\begin{center}
\begin{tabular}{|l|}
\hline
\rowcolor[HTML]{C0C0C0} 
\textbf{User Prompt} \\ 
\hline
\rowcolor[HTML]{EFEFEF} 
\begin{tabular}[c]{@{}l@{}}Describe the video by detailing the following aspects: \\ 
1. The main content and theme of the video.\\ 
2. The color, shape, size, texture, quantity, text, and spatial relationships of the objects.\\ 
3. Actions, events, behaviors temporal relationships, physical movement changes of the objects.\\ 
4. Background environment, light, style and atmosphere.\\ 
5. Camera angles, movements, and transitions used in the video.\\  
Please provide the output in a non-structured way and keep it in one line: \{text\}
\end{tabular} \\ 
\hline
\end{tabular}
\end{center}
\label{tab:user_prompt}
\end{table*}

Inspired by HunyuanVideo, we have designed the following user prompt. Our prompt considers multiple dimensions in a hierarchical manner, which significantly enhances the richness of the output details. Refer to Tab.~\ref{tab:user_prompt} for the specific prompt

\textcolor{white}{In the main text, we presented two examples comparing Dynamic-I2V and CogVideoX-I2V-5B. Due to space constraints, we did not include the results of DynamiCrafter and SVD which are also used in the Human Study. Here, we will supplement the results of all these models. Additionally, extracting frames from a video and displaying them often fails to effectively demonstrate the generated video's quality. On one hand, limited image resolution makes it difficult to showcase certain details. On the other hand, aspects like camera panning, especially in landscape scenes, are better conveyed through video format. Therefore, we have included an additional file, results.zip, which contains several results on our VBench-I2V test set.}

{\small
\bibliographystyle{ieeenat_fullname}
}

%% file: _constants.tex
\def\paperID{9796}
\def\confName{ICCV}
\def\confYear{2025}

\newif\ifreview 
\newif\ifarxiv \newcommand{\arxiv}{\arxivtrue}
\newif\ifcamera 
\newif\ifrebuttal 

%% file: cvpr_header.tex
\ifreview \usepackage[review]{cvpr} \fi
\ifarxiv \usepackage[pagenumbers]{cvpr} \fi
\ifrebuttal \usepackage[rebuttal]{cvpr} \fi
\ifcamera \usepackage{cvpr} \fi

\input{_macros}  % Add packages to _macros.tex

\usepackage{xr-hyper}

\makeatletter
\newcommand*{\addFileDependency}[1]{
  \typeout{(#1)}
  \@addtofilelist{#1}
  \IfFileExists{#1}{}{\typeout{No file #1.}}
}

\makeatother
\newcommand*{\myexternaldocument}[1]{
    \externaldocument{#1}
    \addFileDependency{#1.tex}
    \addFileDependency{#1.aux}
}

\definecolor{cvprblue}{rgb}{0.21,0.49,0.74}
\usepackage[pagebackref,breaklinks,colorlinks,allcolors=cvprblue]{hyperref}
\usepackage[capitalize]{cleveref}
\crefname{section}{Sec.}{Secs.}
\crefname{table}{Table}{Tables}
\crefname{figure}{Fig.}{Figs.}

\ifarxiv \crefname{appendix}{App.}{Apps.}
\else \crefname{appendix}{Suppl.}{Suppls.} \fi

%% file: _macros.tex
\usepackage{graphicx}	
\usepackage{amsmath}	
\usepackage{amssymb}	
\usepackage{booktabs}
\usepackage{times}
\usepackage{microtype}
\usepackage{epsfig}
\usepackage{caption}
\usepackage{float}
\usepackage{placeins}
\usepackage{color, colortbl}
\usepackage{stfloats}
\usepackage{enumitem}
\usepackage{tabularx}
\usepackage{xstring}
\usepackage{multirow}
\usepackage{xspace}
\usepackage{url}
\usepackage{subcaption}
\usepackage{xcolor}
\usepackage[hang,flushmargin]{footmisc}
\usepackage{makecell}

\ifcamera \usepackage[accsupp]{axessibility} \fi

% ...

\ifarxiv  \fi

\newcommand{\R}[1]{{%
    \textbf{%
        \ifstrequal{#1}{1}{\textcolor{red}{R#1}}{%
        \ifstrequal{#1}{2}{\textcolor{blue}{R#1}}{%
        \ifstrequal{#1}{3}{\textcolor{magenta}{R#1}}{%
        \ifstrequal{#1}{4}{\textcolor{teal}{R#1}}{%
                           \textcolor{cyan}{R#1}%
        }}}}%
    }%
}}

%% file: 00_abstract.tex
\begin{abstract}

Recent advancements in image-to-video (I2V) generation have shown promising performance in conventional scenarios. However, these methods still encounter significant challenges when dealing with complex scenes that require a deep understanding of nuanced motion and intricate object-action relationships.
To address these challenges, we present Dynamic-I2V, an innovative framework that integrates Multimodal Large Language Models (MLLMs) to jointly encode visual and textual conditions for a diffusion transformer (DiT) architecture. By leveraging the advanced multimodal understanding capabilities of MLLMs, our model significantly improves motion controllability and temporal coherence in synthesized videos. The inherent multimodality of Dynamic-I2V further enables flexible support for diverse conditional inputs, extending its applicability to various downstream generation tasks. Through systematic analysis, we identify a critical limitation in current I2V benchmarks: a significant bias towards favoring low-dynamic videos, stemming from an inadequate balance between motion complexity and visual quality metrics. To resolve this evaluation gap, we propose DIVE - a novel assessment benchmark specifically designed for comprehensive dynamic quality measurement in I2V generation. In conclusion, extensive quantitative and qualitative experiments confirm that Dynamic-I2V attains state-of-the-art performance in image-to-video generation, particularly revealing significant improvements of 42.5\%, 7.9\%, and 11.8\% in dynamic range, controllability, and quality, respectively, as assessed by the DIVE metric in comparison to existing methods.

% Extensive experiments demonstrate that Dynamic-I2V outperforms existing state-of-the-art methods across both quantitative metrics and human evaluations, while our proposed DIVE benchmark shows xx\% higher sensitivity to motion dynamics compared to conventional evaluation approaches.

\end{abstract}

%% file: 01_intro.tex
\section{Introduction}
\label{sec:intro}

With the advancement of diffusion models\cite{ho2020denoisingdiffusionprobabilisticmodels}, text-to-image (T2I) generation has recently achieved remarkable progress. By integrating temporal information, text-to-video (T2V) generation has also gradually matured. Building upon T2V, image-to-video (I2V) generation seeks to create videos that exhibit dynamic and natural motion while preserving the content of the input image.

\begin{figure}[ht] 
  \centering
  \includegraphics[width=0.5\textwidth]{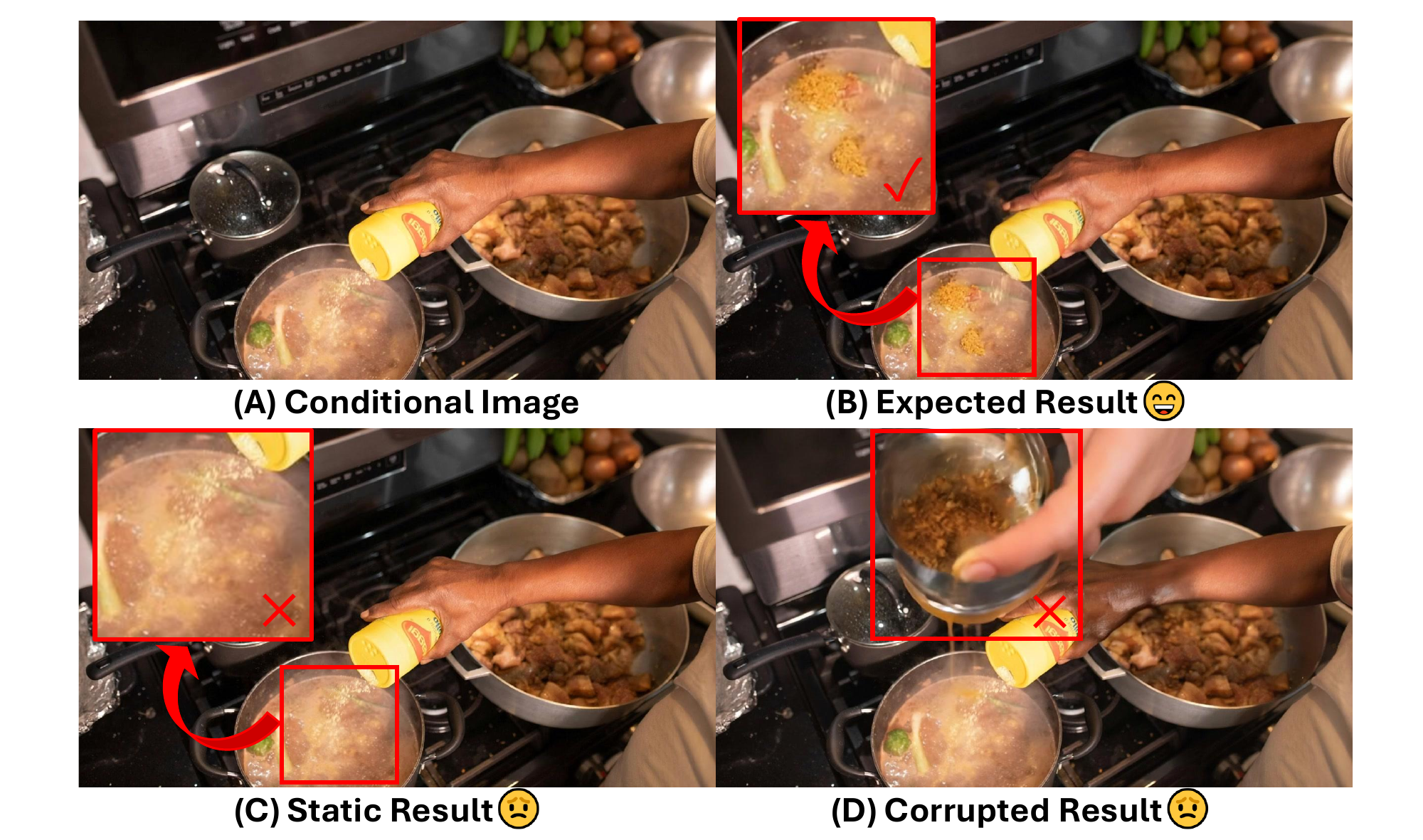}
  \caption{I2V difficulty examples and showcases. (C) generates static video. (D) does not make use of the conditional image object but regenerates the new object. Results show the differences in the ability of various I2V models to understand conditional images.}
  \label{fig:showcase2}
\end{figure}

It has been observed that current I2V models consistently struggle to understand the relationship between instructions and images, often resulting in generated videos that are either static or lack coherent motion. The primary reason for this phenomenon is that current methods lack effective mechanisms for understanding image and text information. For example,  consider the scenario depicted in Fig.~\ref{fig:showcase2}, where the image shows a person holding a spice jar and pouring seasoning into a pot of food, accompanied by the prompt "a person pouring seasoning into a pot of food."  Some methods fail to accurately recognize the spice jar in the person's hand, leading either to the generation of static frames or to the appearance of an extraneous hand with a spice jar to fulfill the action described. Therefore, a key challenge of the I2V task is to: \textbf{Effectively explore and understand the complex relationships between image and text conditions to generate videos that exhibit both dynamism and high quality}

% \textbf{To explore the relationship between complex image and text conditions in order to generate videos that maintain both dynamism and quality.}
%  \textbf{Understanding complex image and text conditions and generating videos that ensure both dynamism and quality.}

To address this challenge, some I2V methods\cite{lei2024animateanythingconsistentcontrollableanimation, zhang2023i2vgenxlhighqualityimagetovideosynthesis, xu2024easyanimatehighperformancelongvideo} leverage vision encoders to extract image features, which are integrated with textual information to enhance video generation quality. Additionally, some approaches incorporate Large Language Models (LLMs) to improve the extraction and comprehension of textual information, thus augmenting the system's capacity to understand and execute instructions. Despite these advancements, current I2V methodologies are still limited in their ability to fully capture the nuances of both imagery and text, particularly in complex scenarios that demand high-quality, dynamic video production.

% Recent multimodal LLMs (MLLMs) have established robust multimodal understanding capabilities, such as accurately interpreting and generating descriptive text for images, performing visual question answering and enabling zero-shot classification based on text prompts. Leveraging powerful pre-trained prior knowledge, MLLMs effectively integrate nuanced visual and natural language information extracted from text and image conditions. 

Recent advancements in multimodal LLMs (MLLMs) have demonstrated significant capabilities in understanding and processing multimodal data. These sophisticated models are now adept at accurately interpreting and generating descriptive text for images, performing complex visual question answering tasks, and enabling zero-shot classification based on textual prompts. 
To enhance the multimodal understanding capability of the I2V model, we leverage the powerful pre-trained priors from MLLM to effectively integrate fine-grained visual information and natural language information extracted from images and texts. In dealing with the diverse features extracted from textual and visual data, we designed a lightweight and learnable multimodal conditional adapter to better fuse these multimodal features.

With the continuous evolution of video generation technology, assessing the quality of generated videos has become increasingly challenging.
Firstly, current metrics primarily focus on the visual quality of individual frames, often overlooking the dynamic aspects of the video. This oversight results in evaluation outcomes that frequently diverge from human subjective judgments. Secondly, there is a notable negative correlation between perceived video quality and the level of motion dynamics in generated videos. As a consequence, models are incentivized to produce videos with reduced motion dynamics in order to achieve higher quality scores on existing benchmarks. 
% An extreme example of this issue is that when we replicate a single image from the VBench-I2V~\cite{huang2024vbenchcomprehensiveversatilebenchmark} test set 49 times to create a completely static video, this method yields overall test scores comparable to the current state-of-the-art performance. The detailed results are presented in Tab.~\ref{tab:our_test}. 
An extreme example is that we used images from the VBench-I2V\cite{huang2024vbenchcomprehensiveversatilebenchmark} test set (the most widely used I2V benchmark) and replicated them to create a completely static video consisting of 49 frames. 
Such videos achieved scores on VBench-I2V comparable to the state-of-the-art method. The detailed results are presented in Tab.~\ref{tab:our_test}. This is contrary to the expectation that dynamic content is a crucial element of video generation. 

% Existing metrics for image-to-video generation often assign a separate score to the degree of motion in videos, which is then incorporated into the overall score with a certain weight. This approach leads to a situation where other metrics are calculated without considering the video's motion level as a prerequisite, resulting in a relatively low weight for motion dynamics in the final score. Moreover, the quality of a video is often inversely correlated with its degree of motion, which can cause relatively static videos to achieve higher scores on benchmarks like VBench~\cite{huang2023vbench}. This discrepancy leads to a misalignment between VBench results and human subjective evaluations. 
% % An extreme example of this issue is that when we replicate a single image from the VBench-I2V~\cite{huang2024vbenchcomprehensiveversatilebenchmark} test set 49 times to create a completely static video, such a video can achieve a score comparable to that of CogVideoX-5B-I2V on VBench.
% An extreme example of this issue is that when we replicate a single image from the VBench-I2V~\cite{huang2024vbenchcomprehensiveversatilebenchmark} test set 49 times to create a completely static video, this method yields overall test scores comparable to the current state-of-the-art performance. The detailed results are presented in Table~\ref{tab:our_test}.

We argue that motion dynamics is the most critical dimension of video quality, and the evaluation of video quality should be based on the degree of motion, encouraging models to generate videos with higher motion dynamics. Based on this principle, we propose DIVE (Dynamics Image-to-Video Evaluation), a metric designed to measure the dynamic range, motion controllability, and motion-based quality of generated videos. DIVE places greater emphasis on encouraging models to produce videos with higher motion dynamics and prioritizes the evaluation of motion levels in videos. Furthermore, its evaluation results exhibit a higher degree of consistency with human subjective assessments.

In summary, our key contributions are as follows.
\begin{itemize}
% \item We introduce Dynamic-I2V, an I2V framework that integrates MLLM and DiT. We have designed an end-to-end architecture that incorporates MLLM into I2V and have thoroughly explored how to better leverage MLLM to extract both image and text information.
\item We propose a novel I2V model, termed Dynamic-I2V, significantly enhancing the text-image understanding capabilities of I2V models by learning dynamic and discriminative feature from MLLMs.
% We propose a novel I2V model, termed Dynamic-I2V, significantly 
The multimodal conditional adapter we designed demonstrates effective multimodal conditional integration. 
% \yanhao{this argument makes MLLM and DiT framework seem too simple, dynamic and other advantages are not leveraged}

\item We demonstrate that leveraging MLLMs endows the I2V model with the capability to input a variety of image conditions, resulting in the generation of videos that are consistent with the semantics and attributes of the conditional images, such as texture and patterns. 

% \item We first investigate that conventional I2V metrics fail to measure video dynamic quality,  and    propose a novel I2V evaluation method, DIVE, which reliably quantifies the dynamic quality of the generated videos. With DIVE metric, our results demonstrate improvements of 42.5\%, 7.9\%, and 11.8\% in dynamic range, controllability, and quality, respectively, when compared to the state-of-the-art methods \cite{yang2024cogvideox}. 

\item We first identify the shortcomings of conventional I2V metrics in assessing the dynamic quality of generated videos. To address this, we propose a novel evaluation method called DIVE, which effectively quantifies the dynamic aspects of video quality. Utilizing the DIVE metric, our approach shows significant improvements over state-of-the-art methods\cite{yang2024cogvideox}, with enhancements of 42.5\% in dynamic range, 7.9\% in dynamic controllability, and 11.8\% in dynamics-based quality.

% \yanhao{what insight are in this metric are not proposed.}
% 更好的capture动态幅度和质量 comparing 传统的I2V benchmark   

\end{itemize}

%% file: 02_related.tex
\section{Related Work}
\label{sec:related}

\subsection{Diffusions with MLLM}

% Diffusion models have gained prominence as effective generative models, starting in the domain of text-to-image generation\cite{GLIDE,ramesh2022hierarchicaltextconditionalimagegeneration,Rombach_Blattmann_Lorenz_Esser_Ommer_2022,Imagen}. They operate by gradually transforming Gaussian noise into coherent data through a denoising process. Some key frameworks, such as Denoising Diffusion Probabilistic Models (DDPM)\cite{ho2020denoisingdiffusionprobabilisticmodels}, demonstrated their capacity to generate high-fidelity images that often surpass traditional methods like GANs\cite{goodfellow2014generativeadversarialnetworks}.
% Building on success in image generation, diffusion models have been adapted for video generation\cite{MakeAVideo}, capturing both spatial and temporal features.  I2V models\cite{yang2024cogvideox, xing2023dynamicrafteranimatingopendomainimages, lin2024stivscalabletextimage, lei2024animateanythingconsistentcontrollableanimation, chen2023seineshorttolongvideodiffusion, zhang2023i2vgenxlhighqualityimagetovideosynthesis, ren2024consisti2venhancingvisualconsistency} represent a specific method for generating dynamic video sequences from static images. 

With the advancement of diffusions and LLMs\cite{qwen, touvron2023llamaopenefficientfoundation, brown2020languagemodelsfewshotlearners, du2022glmgenerallanguagemodel}, several generative methods that integrate LLM and diffusion have emerged. In visual generation and editing tasks, a series of studies\cite{fu2024guidinginstructionbasedimageediting, erwold-2024-qwen2vl-flux, kosmos-g, huang2024smartedit} have adeptly integrated LLM with diffusion models, resulting in notable advancements. 
Particularly, \cite{tan2024mimirimprovingvideodiffusion} achieved good results on T2V by obtaining more detailed text and instruction information using LLM.
Furthermore, Multimodal LLM(MLLMs)\cite{wang2024qwen2vl, wang2024cogvlmvisualexpertpretrained, chen2024internvl, lin2024videollavalearningunitedvisual} can provide more multimodal information compared to LLMs.
Studies\cite{tian2025migeunifiedframeworkmultimodal,sun2024generativemultimodalmodelsincontext,pan2024kosmosggeneratingimagescontext, huang2023smarteditexploringcomplexinstructionbased, ge2025seedxmultimodalmodelsunified} have leveraged the multimodal reasoning capabilities of MLLMs to achieve exciting advancements.
 % In the context of text-to-video generation, several studies \cite{tan2024mimirimprovingvideodiffusion} have introduced MLLMs as novel feature encoders to provide higher-level semantic information. 
 As the capabilities of these models continue to advance, they are increasingly being incorporated into the pre-training pipeline. For instance, HunyuanVideo \cite{kong2025hunyuanvideosystematicframeworklarge} utilizes MLLMs alongside traditional text encoders like CLIP \cite{clip} and T5 \cite{t5}, offering complementary dimensional information and leveraging the models' ability to bridge textual and visual modalities. In the realm of image-to-video generation, the integration of MLLMs becomes even more crucial for effectively fusing textual and visual information. We have integrated Qwen2VL \cite{wang2024qwen2vl} into CogVideoX-I2V \cite{yang2024cogvideox}, achieving state-of-the-art performance in I2V task.

\subsection{Image-to-video Benchmarks}
In the field of video generation, several benchmarks\cite{liu2023evalcrafter, huang2023vbench, sun2025t2vcompbenchcomprehensivebenchmarkcompositional, rawte2024vibe, 10678444} have emerged for evaluating T2V, but there are relatively fewer benchmarks for I2V.
% The field of image-to-video generation has seen a significant increase in the number of models being developed. However, the evaluation of the quality of generated videos remains a challenge, as there are very few evaluation systems capable of accurately reflecting human subjective perceptions. 
I2V has seen a significant increase in model development, yet evaluating the quality of generated videos remains challenging due to a lack of systems that accurately reflect human subjective perceptions.
VBench\cite{huang2023vbench}, the most widely adopted benchmark in the video generation domain, was initially designed exclusively for text-to-video generation. Subsequently, it was extended to include a test set of image-text pairs and dedicated metrics for image-to-video generation, leading to the development of VBench++\cite{huang2024vbenchcomprehensiveversatilebenchmark}, which supports the evaluation of image-generated videos.

%% file: 03_method.tex
\section{Method}
\label{sec:method}

\begin{figure*}[h] 
  \centering
  \includegraphics[width=0.9\textwidth]{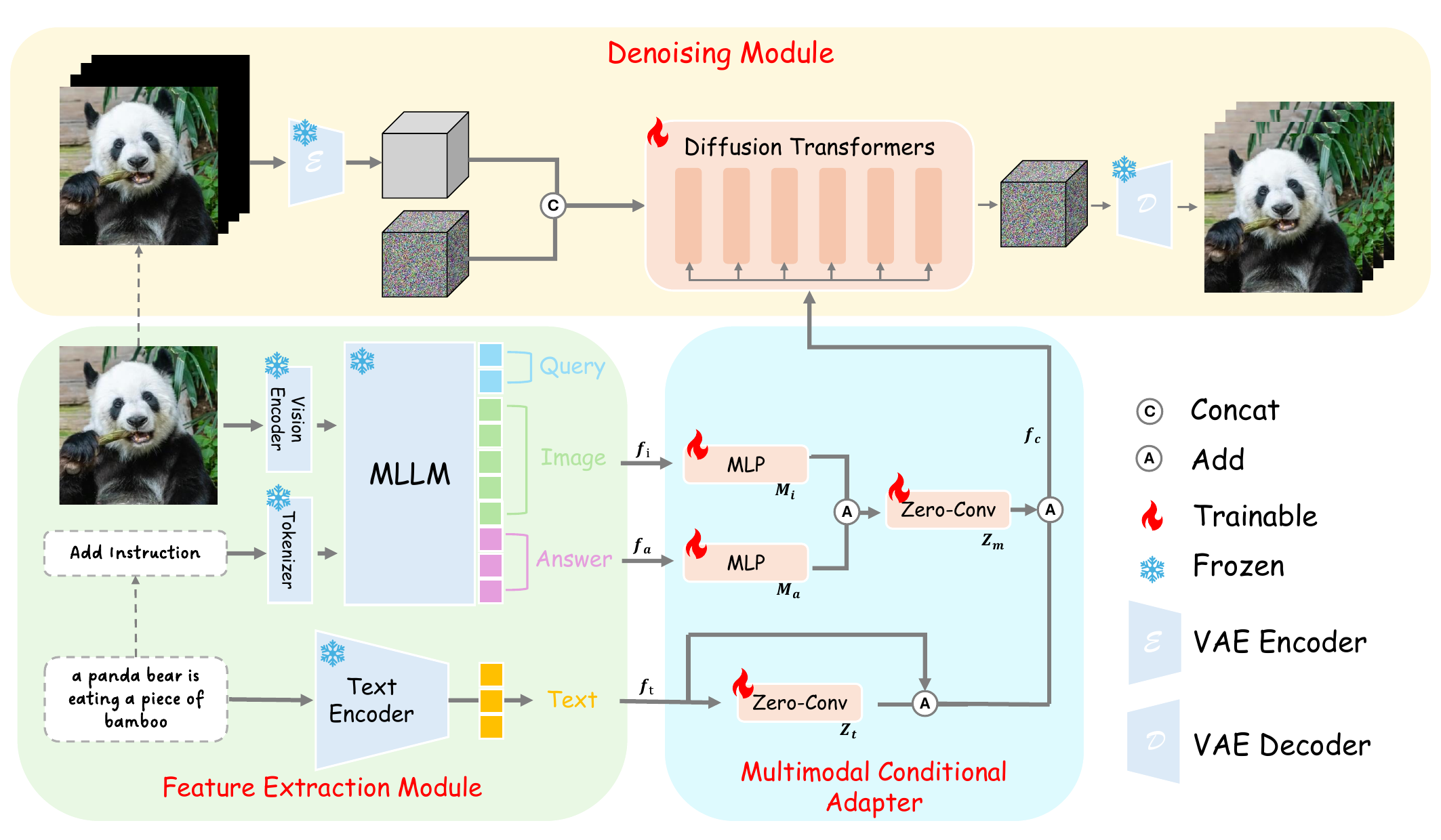}
  \caption{\textbf{Overall of Dynamic-I2V.} Dynamic-I2V consists of a denosing module, a feature extraction module and a multimodal conditional adapter(MCA). MLLM extracts image features \( f_{i}\) and answer features \( f_{a}\) from conditional images and prompts, while the text encoder extracts text features \( f_{t}\). These features, \( f_{i}\), \( f_{a}\) and \( f_{t}\) are then into CA, which integrates them into a comprehensive feature representation \( f_{c}\). This combined feature \( f_{c}\) serves as the final conditional input for the Diffusion Transformers.}
  \label{fig:framework}
\end{figure*}

In this section, we introduce our image-to-video model, Dynamic-I2V, which consists of Dynamic-I2V consists of a denosing module, a feature extraction module and a multimodal conditional adapter(MCA). Afterward, we present the various types of image-conditioned generation supported by Dynamic-I2V. The overview diagram is illustrated in Fig.~\ref{fig:framework}. 

% \begin{figure}[h] 
%   \centering
%   \includegraphics[width=0.45\textwidth]{figs/model3.pdf}
%   \caption{Overall of Dynamic-I2V. Dynamic-I2V consists of a MLLM, a basic video generation model, and a multimodal conditional adapter.}
%   \label{fig:framework}
% \end{figure}

\subsection{Preliminary}
\label{sec:Preliminary}
\subsubsection{Diffusion Models for Image-to-Video Generation} 

% Diffusion-based image-to-video models typically consist of a text encoder, a VAE and a diffusion model.These methods typically use the first frame as condition, concatenated with the initial noise map, as the input to the diffusion model.

Diffusion-based image-to-video models typically consist of a encoder  \(\mathcal{E}\), a decoder\(\mathcal{D} \), a pretrained text encoder \(\mathcal{T} \) and a denoiser \(\epsilon _{\theta }\) .
% These methods typically use the first frame as condition, concatenated with the initial noise map, as the input to the diffusion model.
The initial frame image is frequently utilized as a conditional input, which is then combined with the initial noise map to serve as the primary input for the model.

% Specifically, for a input video, the VAE encoder yields video latents representation of size \([b, f, c, h, w]\) with the compression of both temporal and spatial dimensions.  
Specifically, for the RGB-level condition image \(I\), we first add a small Gaussian noise \( \mathcal{N} (-3.0, 0.5^2) \). 
% The noisy image combine with f-1 frames consisting entirely of zeros to create a new f-frame video, with the noisy image as the first frame. This video, after being encoded by a VAE, yields a latent representation with the same shape of \([b, f, c, h, w]\).
The noisy image, combined with \(f-1\) frames of all zeros, generates a pseudo \(f\)-frame video, with the noisy image as the first frame. This video is then encoded using a encoder \( \mathcal{E}\), resulting in a latent representation \( z_{p} \in \mathbb{R}^{F \times C \times H \times W} \), with temporal and spatial compression.
% The noise-infused image is processed through the VAE to obtain a latent representation with a shape of \([b, 1, c, h, w]\), which is concatenated with a latent representation of size \([b, f-1, c, h, w]\) filled with zeros, resulting in a condition latent of shape \([b, f, c, h, w]\). 
Ultimately, this latent representation is concatenated along the channel dimension with that of the initial video latent, resulting in a model input \( z_{0} \in \mathbb{R}^{F \times 2C \times H \times W} \). Accordingly, the input dimensions for the DiT are also adjusted to match this new input configuration.

During the training phase, provided an input image \textit{I} and a textual condition \(c_{t}\) , the image latent \(z_0 = \varepsilon (I)\) undergoes diffusion through a deterministic Gaussian process over \textit{T} time steps, resulting in the noisy latent representation \(z_T \sim \mathcal{N}(0, 1)\). The training process is focused on mastering the reverse denoising procedure, governed by the following objective function:
\begin{equation}
\mathcal{L} = \mathbb{E}_{\varepsilon (I), \epsilon \sim \mathcal{N}(0,1),t,c_{t}}[\left \| \epsilon -\epsilon _{\theta }(z_{t},t, c_{t}) \right \| _{2}^{2}]
\label{eq:equation1}
\end{equation}
where \( \epsilon _{\theta }\) is the trainable module, \( t = \left \{  1,\dots ,T\right \}  \).

% In our method, this fundamental approach to introducing image conditions will also be utilized.

\subsubsection{MLLMs} 
Multimodal Language Models (MLLMs) are designed to seamlessly integrate visual and textual information, thereby enhancing comprehension and performance in generation tasks. By utilizing a joint encoding framework, it effectively processes images and text, with particular strengths in applications such as image captioning and visual question answering. Its architecture aligns visual features with linguistic representations, thereby enabling the generation of coherent and contextually relevant outputs, and enhancing interactions in multimodal applications.

MLLMs utilize a multimodal input format that amalgamates visual and textual data during processing. When provided an image and its corresponding text as input, the model leverages the contextual information offered by both modalities during the encoding process. In terms of output, MLLMs primarily produce Query Tokens, Vision Tokens, and Answer Tokens. Query Tokens encode the semantic information of the question, while Vision Tokens capture the features of the visual content. The Answer Token represents the model's generated response based on the textual and visual information, ultimately yielding a textual representation as the final answer. In the task of processing multimodal image-text information using MLLM, different output tokens play different roles.

\subsubsection{Benchmark}

% In the field of T2V, a new benchmark named DEVIL\cite{liao2024evaluation} has proposed, which more focuses on dynamics of generated videos. 
% For each prompt, use GPT4o to score its dynamic degree($G$), ranging from 1 to 5. For the generated videos, score the dynamics of each video, ranging from 0 to 1, named dynamic score($S$). Then, three metrics are calculated based on the dynamic degree and dynamic score of each video: dynamic range; dynamic controllability; dynamic-based quality.

In the domain of T2V, a novel benchmark named DEVIL\cite{liao2024evaluation} has been introduced, placing a greater emphasis on the dynamics of the generated videos. For each given prompt, GPT-4o is utilized to score its dynamic degree, which ranges from 1 to 5. 
Subsequently, the dynamics of the corresponding generated videos are assessed, yielding a dynamic score that varies between 0 and 1. Leveraging both the dynamic degree and dynamic score for each video, three key metrics are computed: dynamic range, dynamic controllability, and dynamic-based quality.

\begin{itemize}
    \item \textbf{Dynamics Range} (\textit{DR}) shows the generative ability of the model. An ideal image-generated video model should be able to generate both relatively static videos and highly dynamic videos.
    \item \textbf{Dynamics Controllability} (\textit{DC}) reflects the ability of the model to accurately follow instructions. For low-dynamic instructions, we hope that the model will generate low-dynamic videos, and vice versa. 
    
    % \item \textbf{Dynamics-based Quality} ($DBQ$) can evaluate video quality based on the degree of dynamics in evaluating video quality. Several dimensions are subdivided: \textbf{M}otion \textbf{S}moothness (MS), \textbf{B}ackground \textbf{C}onsistency (BC), \textbf{S}ubject \textbf{C}onsistency (SC), and \textbf{N}aturalness (Nat). The DEVIL\cite{liao2024evaluation} experiment found that these indicators have a significant negative correlation with the dynamic score of the video, so a new evaluation system is designed. It is able to evaluate video quality taking into account video dynamics, encouraging the model to generate videos with higher dynamics.
    
    \item \textbf{Dynamics-based Quality} (\textit{DBQ}) can evaluate video quality based on the degree of dynamics in evaluating video quality. Several dimensions are subdivided: \textbf{M}otion \textbf{S}moothness (MS), \textbf{B}ackground \textbf{C}onsistency (BC), \textbf{S}ubject \textbf{C}onsistency (SC), and \textbf{N}aturalness (Nat). It is able to evaluate video quality taking into account video dynamics, encouraging the model to generate videos with higher dynamics.
\end{itemize}

Such metrics is specifically designed for the T2V task. Therefore, we extend this method to the I2V domain.

\subsection{Dynamic-I2V}

Current I2V models exhibit suboptimal performance when confronted with challenging text or image conditions, often resulting in the generation of static or distorted videos. This is because only T5 encoder in the existing methods is unable to capture the semantic information in the image, especially the components related to text inputs. 
% To address this challenge, we propose Dynamic-I2V to enhance text encoder's performance to extract combined message of text and image.  
% Since MLLM has shown great success in image understanding given specific instructions, we explore MLLM to extract visual and textual features. \textcolor{red}{TODO}
To address this challenge, we propose Dynamic-I2V, a method that leverages MLLM to extract visual and textual information from conditioned images and texts, thereby enhancing the multimodal understanding capability of the generative model.

For text conditional input \( c_t\), it is processed in two branches: 1) Through the T5 text encoder to obtain text features \(f_t\). 2) By extracting multimodal features along with image condition \( c_i\) and a designed user prompt using MLLM. We designed a user prompt that emphasizes image semantics and temporal action information. Detailed description of the user prompt is presented in Supplementary. As mentioned in Sec.~\ref{sec:Preliminary}, the primary outputs of MLLM consist of Query Tokens, Vision Tokens, and Answer Tokens. We denote the output features of the last transformer layer in MLLM that correspond to these three components as \(f_q\), \(f_i\), and \(f_a\). The mathematical expression is as follows:
\begin{equation}
f_{q}, f_{i}, f_{a} = \mathbf{MLLM}(c_{t}, c_{i})
\label{eq:equation2}
\end{equation}

In order to preserve the original model capacities during the early stages of training and to facilitate a smooth transition to new conditions, we designed Conditional Adapter(CA), a learnable module to integrate the outputs of MLLM and text encoder.
Overall, MCA takes the output feature vectors \( f_i \) and \( f_a \) from MLLM, along with the text features  \( f_t \) from text encoder, as inputs, and produces the combined features \( f_c \) that are then fed into the diffusion model.
In detail, MCA contains two MLPs, \( M_{i} \) and \( M_{a} \) , as well as two zero-initialized convolutional layers, \( Z_m \) for MLLM features and  \( Z_t \) for text features, as shown in Fig.~\ref{fig:framework}.
Mathematically, the computational framework of MCA as follow:
\begin{equation}
f_{c}= Z_{m}(M_{i} (f_{i}) + M_{a} (f_{a})) + (f_{t} + Z_{t}(f_{t}) )
\label{eq:equation3}
\end{equation}

The combined features \( f_c \), rich in multimodal information, are fed into the 3D full-attention module of diffusion transformers to guide the generation process. Combining \eqref{eq:equation1} and \eqref{eq:equation3}, the denoising transformers \( \epsilon  _{\theta }\) is optimized to estimate the introduced noise through the minimization of the following loss function:
\begin{equation}
\mathcal{L} = \mathbb{E}_{\varepsilon (I), \epsilon \sim \mathcal{N}(0,1),t,f_{c}}[\left \| \epsilon -\epsilon _{\theta }(z_{t},t, f_{c}) \right \| _{2}^{2}]
\end{equation}

where \( t = \left \{  1,\dots ,T\right \}  \).

\begin{figure}[] 
  \centering
  \includegraphics[width=0.45\textwidth]{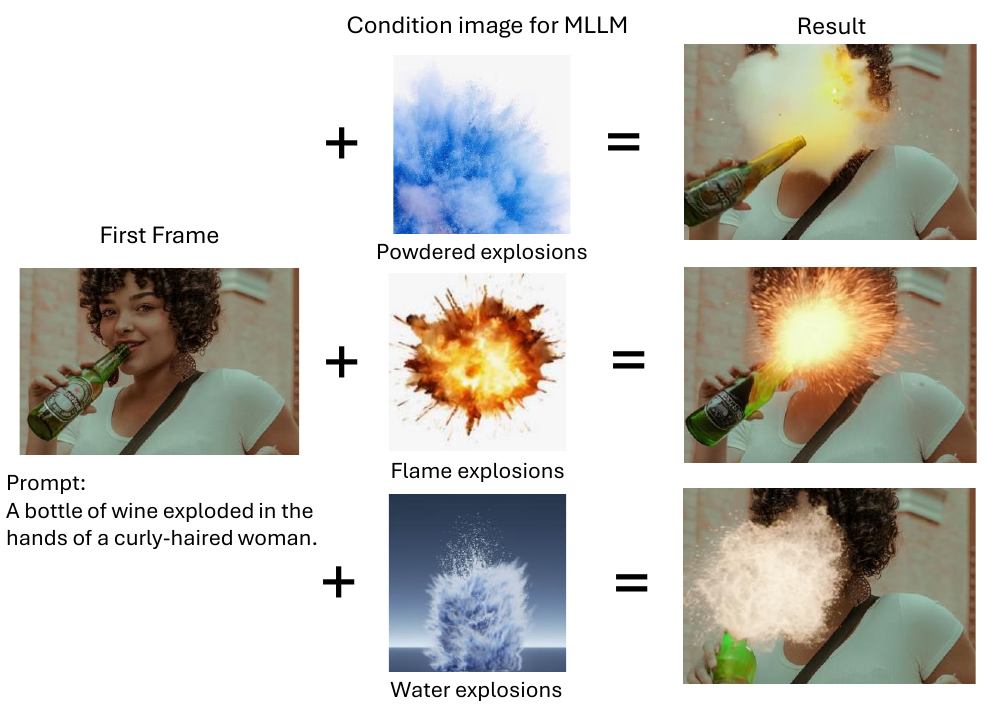}
  \caption{Generation results when using different conditional images for MLLM. The results indicate that the generated video explosion effects vary dynamically according to the explosion type depicted in the conditioning image.}
  \label{fig:condition1}
\end{figure}

\subsection{I2V with other Image Conditions}

% Our methodology benefits from the multimodal understanding capability of MLLM to input more types of image conditions to obtain a variety of higher-order multimodal semantic information. We conducted experiments that revealed a noteworthy observation: when providing a different image to the MLLM, while maintaining all other conditions constant, the generated video exhibits a relationship with the conditioned image. On the one hand, the MLLM can generate Answer Tokens that align with the conditioned image based on the provided picture and adjusted user prompt. On the other hand, the encoding of the conditioned image by the MLLM also offers additional visual contextual information for the generation task.

Our methodology capitalizes on the multimodal comprehension capabilities of MLLM to process a diverse array of image conditions, thereby enabling the extraction of higher-order multimodal semantic information. In our experiments, we revealed a noteworthy observation: altering the input image to the MLLM, while keeping all other parameters constant, influences the generated video's relationship with the conditioning image. Specifically, the MLLM can generate Answer Tokens that are congruent with the conditioned image, as informed by the provided image and the refined user prompt. Moreover, the encoding of the conditioned image by the MLLM contributes supplementary visual contextual information, thereby enhancing the quality and relevance of the generated output.

As illustrated in Fig.~\ref{fig:condition1}, the image features a girl holding a bottle of wine, accompanied by the prompt, "A bottle of wine exploded in the hands of a curly-haired woman." When we presented images depicting various explosion effects to the MLLM, the generated video consistently reflected the specific explosion effect associated with the input image. Experiments demonstrate that MLLM is capable of capturing more diverse semantic information from conditional images to assist in generation. Furthermore, we believe that by constructing specific datasets, MLLM can serve as an effective bridge to provide more complex image conditions to the generation model.

Even without supervised fine-tuning, the results demonstrate that the model can capture abstract semantic information from the given image, which is ultimately reflected in the generated video. We believe that by constructing data and fine-tuning the model, it can attain a stronger capacity for conditional understanding.

%% file: 04_eval.tex
\section{Evaluation Criteria}
\label{sec:eval}

\subsection{VBench-I2V}
The scores from multiple metrics will be subjected to a carefully designed normalization and weighting process to derive the Total Score for VBench-I2V\cite{huang2023vbench, huang2024vbenchcomprehensiveversatilebenchmark}, as depicted in Tab.~\ref{tab:vbenchi2v}. In addition to comparing our results with existing methods on the VBench-I2V benchmark, we also calculated the evaluation results of the CogVideoX-I2V-5B model. It is evident that our model, Dynamic-I2V, has achieved state-of-the-art performance on VBench-I2V with outstanding scores.

% Table
\begin{table*}\small
\caption{\textbf{VBench-I2V Benchmark.} A higher score indicates relatively better performance for a particular dimension. The best and second results for each column are \textbf{bold} and \underline{underlined}, respectively. VBench-I2V includes several metrics as listed below: Video-Image Subject Consistency (\textit{I2V Subj}), Video-Image Background Consistency (\textit{I2V Bkg}), Video-Text Camera Motion (\textit{Cam Mot}), Subject Consistency (\textit{Subj Cons}), Background Consistency (\textit{Bkg Cons}), Motion Smoothness (\textit{Mot Smo}), Dynamic Degree (\textit{Dyn Deg}), Aesthetic Quality (\textit{Aes Qual}), Image Quality (\textit{Img Qual}). CogVideoSFT was implemented by the GaoDeI2V team.}
\begin{tabular}{l|l|l|l|l|l|l|l|l|l|l}
\hline
\textbf{Model} &
  \multicolumn{1}{c|}{\textbf{\begin{tabular}[c]{@{}c@{}}I2V\\ Subj\end{tabular}}} &
  \multicolumn{1}{c|}{\textbf{\begin{tabular}[c]{@{}c@{}}I2V\\ Bkg\end{tabular}}} &
  \multicolumn{1}{c|}{\textbf{\begin{tabular}[c]{@{}c@{}}Cam\\ Mot\end{tabular}}} &
  \multicolumn{1}{c|}{\textbf{\begin{tabular}[c]{@{}c@{}}Subj\\ Cons\end{tabular}}} &
  \multicolumn{1}{c|}{\textbf{\begin{tabular}[c]{@{}c@{}}Bkg\\ Cons\end{tabular}}} &
  \multicolumn{1}{c|}{\textbf{\begin{tabular}[c]{@{}c@{}}Mot\\ Smo\end{tabular}}} &
  \multicolumn{1}{c|}{\textbf{\begin{tabular}[c]{@{}c@{}}Dyn\\ Deg\end{tabular}}} &
  \multicolumn{1}{c|}{\textbf{\begin{tabular}[c]{@{}c@{}}Aes\\ Qual\end{tabular}}} &
  \multicolumn{1}{c|}{\textbf{\begin{tabular}[c]{@{}c@{}}Img\\ Qual\end{tabular}}} &
  \multicolumn{1}{c}{\textbf{\begin{tabular}[c]{@{}c@{}}Total\\ Score\end{tabular}}} \\ \hline
CogVideoSFT\cite{yang2024cogvideox} & 97.67\% & 98.76\% & \underline{84.93}\% & 95.47\% & 98.30\% & 98.35\% & 36.51\% & 59.76\% & 67.64\% & 87.98\% \\
DynamiCrafter-1024\cite{xing2023dynamicrafteranimatingopendomainimages}    & 98.17\% & 98.60\% & 35.81\% & 95.69\% & 97.38\% & 97.38\% & \textbf{47.40}\% & \underline{66.46}\% & 69.34\% & 87.76\% \\
STIV\cite{lin2024stivscalabletextimage}                  & \textbf{98.96}\% & 97.35\% & 11.17\% & \underline{98.40}\% & \underline{98.39}\% & \textbf{99.61}\% & 15.28\% & 66.00\% & 70.81\% & 86.73\% \\
Animate-Anything\cite{lei2024animateanythingconsistentcontrollableanimation}      & 98.76\% & 98.58\% & 13.08\% & \textbf{98.90}\% & 98.19\% & 98.61\% & 2.68\%  & \textbf{67.12}\% & \textbf{72.09}\% & 86.48\% \\
SEINE-512\cite{chen2023seineshorttolongvideodiffusion}        & 97.15\% & 96.94\% & 20.97\% & 95.28\% & 97.12\% & 97.12\% & 27.07\% & 64.55\% & \underline{71.39}\% & 85.52\% \\
I2VGen-XL\cite{zhang2023i2vgenxlhighqualityimagetovideosynthesis}             & 96.48\% & 96.83\% & 18.48\% & 94.18\% & 97.09\% & 98.34\% & 26.10\% & 64.82\% & 69.14\% & 85.28\% \\
ConsistI2V\cite{ren2024consisti2venhancingvisualconsistency}            & 95.82\% & 95.95\% & 33.92\% & 95.27\% & 98.28\% & 97.38\% & 18.62\% & 59.00\% & 66.92\% & 84.07\% \\
SVD-XT-1.1\cite{blattmann2023stablevideodiffusionscaling}            & 97.51\% & 97.62\% & \multicolumn{1}{c|}{-}   & 95.42\% & 96.77\% & 98.12\% & \underline{43.17}\% & 60.23\% & 70.23\% & \multicolumn{1}{c}{-}   \\

CogVideoX-I2V-5B\cite{yang2024cogvideox}      & \underline{98.87}\% & \textbf{99.08}\% & 76.25\% & 96.99\% & \textbf{99.02}\% & 98.85\% & 21.79\% & 60.76\% & 69.53\% & \underline{88.21}\% \\ \hline
Dynamic-I2V           & 98.83\% & \underline{98.97}\% & \textbf{88.10}\%                & 96.21\% & \underline{98.39}\% & \underline{98.88}\% & 27.15\% & 60.10\% & 69.23\% & \textbf{88.45}\%               \\ \hline
\end{tabular}
\label{tab:vbenchi2v}
\end{table*}

% Table
\begin{table}[]\small
\begin{minipage}{0.48\textwidth} %{\columnwidth}
\setlength{\abovecaptionskip}{0pt}
\caption{VBench-I2V without Camera Motion. The best and second results for each column are \textbf{bold} and \underline{underlined}, respectively. Total Score(\textit{TS}), I2V Score(\textit{IS}), Quality Score(\textit{QS})}
\begin{center}
\begin{tabular}{lllll}
  \toprule
    Model & \textit{TS}~$\uparrow$ & \textit{IS}~$\uparrow$ & \textit{QS}~$\uparrow$ \\ \midrule
    Static Videos   & \textbf{88.88}   & \textbf{99.35}   & 78.74  \\ 
    SVD   & 88.06   & 96.79 & \textbf{79.34}  \\ 
    DynamiCrafter   & 88.85   & 98.43 & 79.26  \\ 
    CogVideoX-I2V-5B   & 88.74   & \underline{98.72} & 78.77  \\ 
    Dynamic-I2V   & \underline{88.84}   & 98.36 & \underline{79.32}  \\ 
  \bottomrule
\end{tabular}
\end{center}
\vspace{-20pt}
\label{tab:our_test}
\end{minipage}
\end{table}%

However, our analysis also revealed that VBench exhibits several significant limitations. we selected several open-source models from the VBench-I2V leaderboard for replication and benchmarking, obtaining the evaluation results shown in Tab.~\ref{tab:our_test}. Due to the additional prompt used for testing camera motion while other tests utilized a uniform prompt, we have excluded the evaluation of camera motion.
These results do not align with our subjective perception of the models. To demonstrate the limitations of VBench, we generated purely static videos by duplicating a single image into 49 frames using its test set and conducted evaluations. Surprisingly, these static videos achieved state-of-the-art scores on VBench-I2V without camera motion. This reveals that VBench's scoring system does not sufficiently emphasize the dynamic aspects of videos and exhibits weak consistency with human subjective perception.

\subsection{DIVE}

% Table
\begin{table}[]\small
\begin{minipage}{0.48\textwidth} %{\columnwidth}
\caption{Quantitative comparison on DIVE. \textit{DR}, \textit{DC}, and \textit{DBQ} stand for the three metrics of DIVE: Dynamic Range, Dynamics Controllability, and Dynamics-Based Quality. The best and second results for each column are \textbf{bold} and \underline{underlined}, respectively.}
\vspace{-5pt}
\begin{center}
\begin{tabular}{lllll}
  \toprule
    Model & \textit{DR}~$\uparrow$ & \textit{DC}~$\uparrow$ & \textit{DBQ}~$\uparrow$ \\ \midrule
    % Model & \makecell{Dynamics \\ Range $\uparrow$} & \makecell{Dynamics \\ Controllability $\uparrow$} & \makecell{Dynamics-based \\ Quality $\uparrow$} \\  
    % \midrule
    Static Videos   & 0.89   & 51.03   & 8.26  \\ 
    SVD   & 26.76   & 57.71 & 47.15  \\ 
    DynamiCrafter   & 29.35   & \underline{72.12} & 46.64  \\ 
    CogVideoX-I2V-5B   & \underline{29.61}   & 68.17 & \underline{47.48}  \\ 
    Dynamic-I2V   & \textbf{45.77}   & \textbf{74.44} & \textbf{62.25}  \\ 
  \bottomrule
\end{tabular}
\end{center}
\vspace{-20pt}
\label{tab:dive}
\end{minipage}
\end{table}%

% We use the VBench-I2V test set and DEVIL's evaluation system to form a new image-to-video generation benchmark---DIVE (dynamic image-to-video evaluation). According to DEVIL's approach, we use GPT-4o to score the dynamics of a total of 355 image-text pairs in the VBench I2V test set, ranging from 1 to 5, named dynamic degree($G$), and the score must be an integer. For the 355 generated videos, first score the dynamics of each video, ranging from 0 to 1, named dynamic score($S$). Then, three metrics are calculated based on the dynamic score and dynamic degree of each video: dynamic range; dynamic controllability; dynamic-based quality. The detailed scores are shown in Table~\ref{tab:dynamics_metrics} and Table~\ref{tab:quility_metrics}. The scores for static videos are notably low, which aligns with our subjective observations. Additionally, our model achieves the highest score on the DIVE benchmark, demonstrating its superior performance in generating dynamic and engaging video content. This consistency between objective metrics and subjective evaluation underscores the effectiveness of our approach.
% For more implementation details, refer to DEVIL\cite{liao2024evaluation}.

To better evaluate the dynamic range and content quality of generated videos, we employed the VBench-I2V test set alongside a novel evaluation framework to establish a new benchmark for image-to-video generation, termed DIVE (Dynamic Image-to-Video Evaluation). Following the methodology loosely inspired by DEVIL\cite{liao2024evaluation}, we utilized GPT-4o to assess the dynamic properties within a set of 355 image-text pairs from the VBench-I2V test set. These dynamics were quantified with a dynamic degree, rated on an integer scale from 1 to 5.
For the corresponding 355 generated videos, each was evaluated on its dynamic attributes with a dynamic score, which ranged from 0 to 1. Subsequently, three key metrics were calculated to reflect the dynamic properties and overall quality of the videos: dynamic range, dynamic controllability, and dynamic-based quality. Detailed scores are presented in Tab.~\ref{tab:dive} and Tab.~\ref{tab:dbq}.
Notably, videos with static content received lower scores, echoing our subjective assessments. Our model demonstrated the highest performance on the DIVE benchmark, signifying its capability in creating dynamic and engaging video content. This agreement between the proposed objective metrics and subjective evaluations highlights the efficacy of our methodology.

% Table
\begin{table}[]\small
\begin{minipage}{0.48\textwidth} %{\columnwidth}
\caption{Dynamic-based Quality. \textit{MS}, \textit{BC}, \textit{SC}, and \textit{Nat} represent Motion Smoothness, Background Consistency, Subject Consistency, and Naturalness. The best and second results for each column are highlighted in \textbf{bold} and \underline{underlined}, respectively. Note that SVD generated videos without prompts.}
\setlength{\abovecaptionskip}{0pt}
\vspace{-10pt}
\begin{center}
\begin{tabular}{lllll}
  \toprule
    Model & \textit{MS}~$\uparrow$ & \textit{BC}~$\uparrow$ & \textit{SC}~$\uparrow$ & \textit{Nat}~$\uparrow$ \\ \midrule
    % Static Videos   & 8.31   & 8.25   & 8.31 & 8.15  \\ 
    SVD   & 48.73   & \underline{47.84} & 46.17 & 45.86  \\ 
    DynamiCrafter   & 47.84   & 47.19 & 45.39 & 46.13 \\ 
    CogVideoX-I2V-5B   & \underline{49.24}   & 46.44 & \underline{46.30} & \underline{47.95} \\ 
    Dynamic-I2V   & \textbf{65.26}   & \textbf{62.07} & \textbf{59.00} & \textbf{62.66} \\ 
  \bottomrule
\end{tabular}
\end{center}
\vspace{-20pt}
\label{tab:dbq}
\end{minipage}
\end{table}%

\subsection{Human Study}
We engaged 20 volunteers with expertise in video generation to conduct a human evaluation of the four representative models: CogVideoX-I2V-5B, DynamiCrafter, SVD and our Dynamic-I2V. The detailed guidelines for this manual assessment are provided in supplementary.

We randomly selected 50 videos from the test set and asked the volunteers to rank results from the four test models based on different criteria: video dynamics, video naturalness, and video text compliance. Finally, they also need to provide an overall quality score for each video. Points were assigned based on the ranking, with 4 points for the first place, 3 points for the second, 2 points for the third, and 1 point for the fourth. This allowed us to derive scores for the three specific dimensions and the overall quality. The proportion of scores for each model was then calculated. The results of this evaluation are presented in Tab.~\ref{tab:human_evaluation} and an example is shown in the Fig.~\ref{fig:Examples}.

The results indicate that both the DIVE and human study are ranked in the following order: Dynamic-I2V \textgreater CogVideoX-I2V-5B \textgreater DynamiCrafter \textgreater SVD. However, VBench presents a different order. This indicates that DIVE is closer to human subjective perception compared to VBench.

% Among the four selected models, the results of our proposed DIVE align consistently with the subjective evaluation outcomes, whereas the results from VBench exhibit discrepancies with the subjective evaluation. The specific sorting is Dynamic-I2V \textgreater CogVideoX-I2V-5B \textgreater DynamiCrafter \textgreater SVD. 

% As illustrated in Table ~\ref{tab:Rank}.

% \begin{table}[H]\small
% \begin{minipage}{0.48\textwidth} %{\columnwidth}
% \caption{Model performance ranking of three evaluation methods. Rankings that differ from human evaluation are marked in \textcolor{red}{red}.}
% \begin{center}
% \begin{tabular}{lccc}
%   \toprule
%     Model & Human Eval & DIVE & Vbench  \\ \midrule
%     SVD & 4   & 4 & 4\\ 
%     DynamiCrafter    & 3   & 3 & \textcolor{red}{2} \\ 
%     CogVideoX-I2V-5B   & 2   & 2 & \textcolor{red}{3} \\ 
%     Dynamic-I2V   & 1   & 1 & 1 \\ 
%   \bottomrule
% \end{tabular}
% \end{center}
% \label{tab:Rank}
% % \bigskip\centering
% \end{minipage}
% \end{table}

%% file: 05_exp.tex
\section{Experiments}
\label{sec:exp}

\subsection{Data Preprocess}
% Data Preprocess
In our experiments, we utilize the open-source dataset OpenVid \cite{nan2024openvid1mlargescalehighqualitydataset}. Through our research, we identified several advantages and disadvantages of the OpenVid dataset. A notable strength is its inclusion of the 'camera motion' field, which aligns closely with the metrics used in VBench. This specific feature is not present in other current open-source datasets\cite{Bain21, ju2024miradatalargescalevideodataset, wang2024koala36mlargescalevideodataset, chen2024panda70m}. For the dataset consisting of 1 million videos, we first filtered out those with uncertain and mixed camera motion, resulting in approximately 350k videos remaining.
Based on our meticulous observation, the remaining videos still exhibit some imperfections. Firstly, many of these remaining videos still contain transition animations, which can affect video quality by introducing transitions in the inference results. To address this issue, we employed CoTracker~\cite{karaev2024cotrackerbettertrack} to verify the single camera motion types and detect transition problems. After this additional filtering, the size of the OpenVid dataset was further reduced to 123k.

% Table
\begin{table}[]\small
\begin{minipage}{0.48\textwidth} %{\columnwidth}
\setlength{\abovecaptionskip}{3pt}
\caption{Human Evaluation: Calculate the score proportion of each model. The best and second results for each column are \textbf{bold} and \underline{underlined}, respectively. The four dimensions are Dynamics(\textit{D}), Naturalness(\textit{N}), Text Conpliance(\textit{TC}), overall quality(\textit{All})}
\begin{center}
\begin{tabular}{lllll}
  \toprule
    Model & \textit{D}~$\uparrow$ & \textit{N}~$\uparrow$ & \textit{TC}~$\uparrow$ & \textit{All}~$\uparrow$ \\ \midrule
    SVD & \underline{24.15}   & 17.19 & 18.35 & 16.96 \\ 
    DynamiCrafter    & 23.56   & 21.96 & 21.72 & 21.98   \\ 
    CogVideoX-I2V-5B   & 22.33   & \underline{28.39} & \underline{27.15} & \underline{28.11}  \\ 
    Dynamic-I2V   & \textbf{29.96}   & \textbf{32.46} & \textbf{32.77} & \textbf{32.95}  \\ 
  \bottomrule
\end{tabular}
\end{center}
\vspace{-20pt}
\label{tab:human_evaluation}
\end{minipage}
\end{table}%

\begin{figure*}[] 
  \centering
  \includegraphics[width=0.88\textwidth]{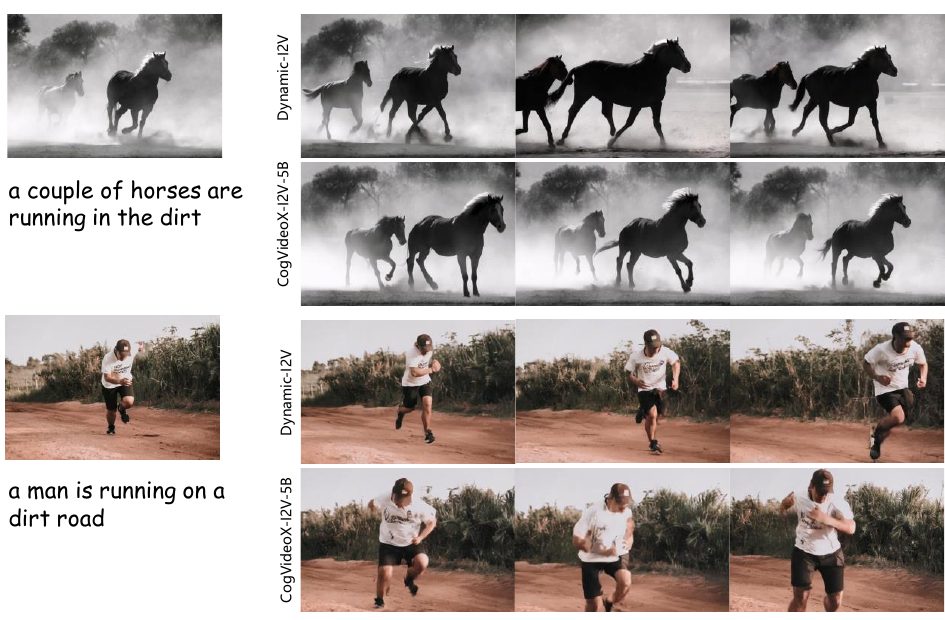}
  \caption{Subjective comparison between CogVideoX-5B-I2V\cite{yang2024cogvideox} 
 with our Dynamic-I2V. The first group demonstrates the stronger dynamic amplitude of Dynamic-I2V, where the comparison horse has a smaller movement distance while our horse runs forward more extensively. The second group showcases our results in maintaining the quality of the moving subject better amidst larger subject movements. }
 \vspace{-5pt}
\label{fig:Examples}
\end{figure*}

\subsection{Implementation Details}

We convert all video data to a resolution \(480 \times 720\), with a maximum frame length of 49. During data collection, we dynamically adjust the frame sampling interval based on the video's duration to ensure that we capture 49 frames evenly distributed throughout the entire video.
Our base model for video generation is CogVideoX-I2V-5B, which we initialize following the fine-tuning configuration used in the SAT setup of CogVideoX. For MLLM, we employ the Qwen2VL-2B model. Considering GPU memory limitations, our batch size is set to 1. We utilize 16 GPUs for fine-tuning on 123k videos from OpenVid. With an iteration time of 18 seconds, each epoch for this dataset takes approximately 38 hours. Our learning rate is set to 0.00001, and we utilize the FusedEmaAdam optimizer with betas of [0.9, 0.95].

\subsection{Ablation Study}

We initiated our ablation study with the baseline model of CogVideoX-I2V-5B, maintaining a consistent training configuration throughout the investigation.
To enhance the assessment of dynamism and visual quality in the generated videos, we employed the DIVE evaluation method for these experiments. Results are presented in Tab.~\ref{tab:ablation_study}.

\textbf{Baseline + MLLM - T5} indicates that extracting multimodal features directly using MLLM, instead of utilizing T5 text features, results reveal a deterioration of DIVE scores compared to the baseline. This decline is likely attributable to the substantial amount of data required for adaptation to the direct replacement.

\textbf{Baseline + MLLM} denotes the simultaneous use of MLLM while retaining T5, combining the features of both as conditions through simple addition. The results indicate that retaining T5 effectively ensures overall performance. However, there is no significant improvement compared to the baseline.

% The results illustrate a slight increase in the VBench values; however, they remain lower than the baseline. This decline can primarily be attributed to a decrease in dynamicity, despite improvements in subject consistency and image quality.

\textbf{Baseline + MLLM + MLPs} refers to retaining T5 while integrating MLLM, where each MLLM feature is processed through a learnable MLP and then combined via addition. Experimental results demonstrate that re-learning the features of MLLM through a trainable MLP can enhance the utilization of extracted conditional information for generation purposes.

\textbf{Baseline + MLLM + MCA (Ours)} indicates that our final structure, Dynamic-I2V, leverages both MLLM and a text encoder to extract visual and textual features, and subsequently employs a Multimodal Conditional Adapter(MCA) module for feature fusion, achieving optimal performance across the metrics Dynamic Range, Dynamics Controllability, and Dynamics-Based Quality.

% we not only utilized the output features from MLLM and retained T5 but also employed Conditional Adapter(CA) to replace the direct summation, thereby enhancing compatibility between the two. The VBench results suggest that this approach is very close to the baseline, yet our model demonstrates a significant improvement in dynamicity, surpassing the baseline performance.

% Table
\begin{table}[]\small
\begin{minipage}{\columnwidth}
\caption{Ablation study results on DIVE. \textit{DR}, \textit{DC}, and \textit{DBQ} stand for the three metrics of DIVE: Dynamic Range, Dynamics Controllability, and Dynamics-Based Quality. The best and second results for each column are \textbf{bold} and 
\underline{underlined}, respectively.}
\vspace{-10pt}
\begin{center}
\begin{tabular}{lllll}
  \toprule
    Model & \textit{DR}~$\uparrow$ & \textit{DC}~$\uparrow$ & \textit{DBQ}~$\uparrow$ \\ \midrule
    Baseline   & \underline{29.61}   & 68.17   & 47.48  \\ 
    + MLLM - T5   & 24.63   & 65.81 & 40.05  \\ 
    + MLLM    & 26.88   & 72.62 & \underline{48.29}  \\ 
    + MLLM + MLPs & 27.90 & \underline{72.79} & 47.97 \\ 		
    + MLLM + MCA(ours)   & \textbf{45.77}   & \textbf{74.44} & \textbf{62.25}  \\ 
  \bottomrule
\end{tabular}
\end{center}
\vspace{-20pt}
\label{tab:ablation_study}
\end{minipage}
\end{table}%

%% file: 10_conclusion.tex
\section{Conclusion}
\label{sec:conclusion}

% We propose a image-to-video diffusion model named Dynamic-I2V, which effectively extracts and integrates image and textual information utilizing an MLLM to facilitate a precise understanding of both images and text within the I2V framework. We focus on the optimal utilization of the MLLM, encompassing multiple output modalities and fusion techniques that incorporate various contextual conditions. Furthermore, we introduce a novel evaluation metric for I2V, termed DIVE, which balances the dynamics and quality of the video, thus enhancing the evaluative dimensions of existing I2V assessment methodologies. Our final results demonstrate state-of-the-art performance on both the VBench-I2V and DIVE benchmarks.

We propose an innovative image-to-video diffusion model named Dynamic-I2V, which excels in extracting and integrating both visual and textual information. By leveraging the capabilities of a Multimodal Large Language Model (MLLM), our approach attains a refined understanding of both images and text within the Image-to-Video (I2V) framework. Emphasizing the optimal utilization of MLLM, Dynamic-I2V incorporates multiple output modalities and employs fusion techniques that integrate various contextual conditions.

Moreover, we introduce a novel evaluation metric for the I2V domain, termed DIVE. This metric successfully balances the dynamics and the quality of generated videos, thereby enriching the evaluative dimensions of existing I2V assessment methodologies. Our comprehensive experiments and analyses demonstrate that Dynamic-I2V achieves state-of-the-art performance, outperforming previous methods on both the VBench-I2V and DIVE benchmarks.